%% file: main.tex
\documentclass[letterpaper]{article} 
\usepackage{aaai2026}  
\usepackage{times}  
\usepackage{helvet}  
\usepackage{courier}  
\usepackage[hyphens]{url}  
\usepackage{graphicx} 
\usepackage{natbib}  
\usepackage{caption} 
\usepackage{algorithm}
\usepackage{algorithmic}

\usepackage{amsmath} 
\usepackage{amssymb} 
\usepackage{booktabs}   
\usepackage{multirow} 
\usepackage{makecell}
\usepackage{graphicx}   
\usepackage[table,dvipsnames]{xcolor}
\definecolor{navyblue}{HTML}{0071BC}
\usepackage{subcaption} 
\usepackage{tcolorbox}

\usepackage{newfloat}
\usepackage{listings}
\DeclareCaptionStyle{ruled}{labelfont=normalfont,labelsep=colon,strut=off} 
\floatstyle{ruled}
\newfloat{listing}{tb}{lst}{}
\floatname{listing}{Listing}
\definecolor{MyDarkBlue}{HTML}{1A4B84}
\nocopyright 

\title{SIEVE: Structure-Aware Data Selection for Imitation Learning with VLA Models}

\author{
    Changti Wu\textsuperscript{\rm 1,\rm 2}\thanks{These authors contributed equally},\space
    Bin Yu\textsuperscript{\rm 3,\rm 2}\footnotemark[1],\space
    Zhaolong Shen\textsuperscript{\rm 4,\rm 2}\footnotemark[1],\space
    Shijie Lian\textsuperscript{\rm 5,\rm 2},\space
    Xiaopeng Lin\textsuperscript{\rm 6,\rm 8},\\
    Cong Huang\textsuperscript{\rm 7},\space
    Zhirui Zhang\textsuperscript{\rm 8},\space
    Lei Zhang\textsuperscript{\rm 1}\thanks{Corresponding author},\space
    Kai Chen\textsuperscript{\rm 7,\rm 2,\rm 8}\footnotemark[2] 
}
\affiliations{
    \textsuperscript{1}East China Normal University\space
    \textsuperscript{2}Zhongguancun Academy\space
    \textsuperscript{3}Harbin Institute of Technology\space
    \textsuperscript{4}Beihang University\\
    \textsuperscript{5}Huazhong University of Science and Technology\space
    \textsuperscript{6}The Hong Kong University of Science and Technology (Guangzhou)\space
    \textsuperscript{7}Zhongguancun Institute of Artificial Intelligence\space
    \textsuperscript{8}DeepCybo
}

\begin{document}

\maketitle

\input{sec/0_abstract}

\begin{links}
    \link{Code}{https://github.com/ChangtiWu/SIEVE}
\end{links}

\input{sec/1_intro}
\input{sec/2_related}

\input{sec/3_method}
\input{sec/4_exp}

\input{sec/5_conclusion}

\clearpage
\bibliography{aaai2026}

\input{sec/6_appendix}
\end{document}

%% file: sec/0_abstract.tex
\begin{abstract}
Vision-Language-Action (VLA) models are typically trained by imitation learning on large-scale robot demonstration datasets, but more data does not necessarily yield better policies due to redundancy, noise, and uneven coverage. 
Existing data selection methods often assess demonstrations at either the trajectory or state-action level, missing the reusable structures that compose long-horizon behaviors.
In this paper, we propose SIEVE, a structure-aware data selection method for VLA imitation learning. SIEVE views demonstrations as compositions of reusable primitives and transition interfaces.
It first discovers visuo-motor primitives from segmented trajectories, then allocates selection budgets to composition patterns by maximizing reuse-aware structural exposure under diminishing returns. Finally, it selects medoid trajectories within each composition-pattern bucket to retain central, stable, and imitation-friendly demonstrations.
Experiments across multiple datasets, benchmarks, and VLA models show that SIEVE consistently outperforms competitive data selection baselines. Notably, SIEVE can surpass full-data training while using only 50\% of demonstrations and 50\% of training steps, suggesting that reusable structure, captured through primitives and transitions, is an important signal for efficient VLA imitation learning.
\end{abstract}


%% file: sec/1_intro.tex
\section{Introduction}
Vision-Language-Action (VLA) models have emerged as a scalable paradigm for robotic control, typically acquiring manipulation skills through imitation learning (IL) over large-scale demonstrations \cite{zitkovich2023rt,o2024open,kim2024openvla,black2024pi_0,intelligence2025pi_}. However, the rapid growth of robot demonstration datasets does not automatically translate into better policies. In practice, such datasets often contain substantial trajectory redundancy, noisy human demonstrations, suboptimal behaviors, and uneven task coverage \cite{sathyanarayan2025quality,belkhale2023data,lin2025data,xing2025shortcut}. Training on unfiltered data can repeatedly expose the model to near-duplicate behaviors while also propagating inconsistent or low-quality supervision. These issues make data selection an increasingly important problem for VLA imitation learning: given a large demonstration pool, we aim to retain a compact subset that is more beneficial for policy learning.

\begin{figure}[!t]
  \centering
  \includegraphics[width=1\linewidth]{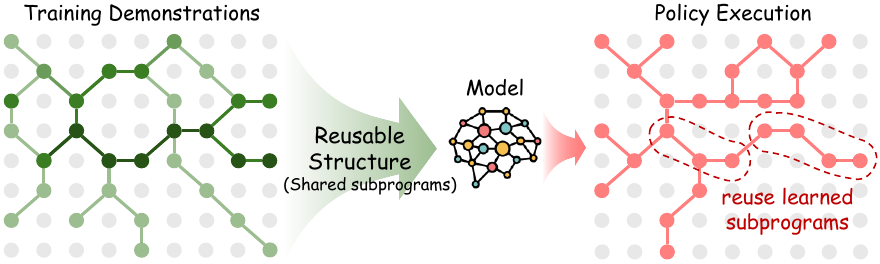}
   \caption{
Motivation of SIEVE. Training demonstrations contain recurring primitive-transition patterns that can be viewed as reusable behavioral subprograms. Inspired by the MDL principle, SIEVE aims to select demonstrations that expose such reusable structures, enabling the policy to internalize shared subprograms.
}
\label{fig:moti}
\end{figure}

Existing data selection methods for imitation learning typically curate demonstrations by estimating sample utility at different granularities. One line relies on trajectory-level signals, such as trajectory-representation similarity for redundancy removal, demonstration reliability, or downstream task feedback \cite{dass2025datamil,chen2025curating,zhang2025scizor,xu2026athena}. While these signals provide a global view of demonstration utility, they may collapse a long-horizon trajectory into a single score and obscure which internal stages or behavior compositions are useful; moreover, feedback-based methods often require costly additional model training. Another line estimates utility from state-action-level signals, such as state-action mutual information, task progress, or joint state-action similarity \cite{zhang2025scizor,hejna2025robot}. These fine-grained signals can capture local predictability or redundancy, but they are often either too myopic to characterize coherent long-horizon task semantics or primarily designed for local pruning. This creates a granularity mismatch: effective IL data selection requires evidence that is coarser than individual state-action pairs, yet more structured than holistic trajectory scores.

Inspired by the Minimum Description Length (MDL) principle \cite{barron1998minimum,rissanen2004minimum,grunwald2007minimum}, we view useful demonstrations as those that expose reusable behavioral regularities. In its two-part form, MDL seeks a description that minimizes the cost of encoding both the model and the data:
\begin{equation}
L(x)=\min_{H\in\mathcal{H}}\left[L(H)-\log P(x\mid H)\right],
\end{equation}
where $L(H)$ is the number of bits required to encode the model $H$, and $-\log P(x\mid H)$ is the number of bits required to encode the data $x$.
This formulation operationalizes Occam's razor: a model is penalized unless it yields a shorter description of the data. When repeating patterns exist, they can be stored in the model rather than redundantly encoded in each data instance.
This view implies that a learner tends to compress data by absorbing repeating patterns into its parameters, so as to encode more data under limited parameters and computation budgets. Useful data, therefore, is not merely abundant or locally predictable, but rich in extractable structure (i.e., non-random regularities that a bounded learner can internalize and reuse as shared behavioral subprograms) \cite{finzi2026entropy}.
For robotic imitation learning, such structure is naturally expressed as primitive composition: a trajectory consists of atomic behavior primitives and transition interfaces. Thus, we estimate trajectory utility by the primitives it exposes, how they are composed, and which transitions support long-horizon execution, yielding a mid-level scoring granularity that captures reusable behavior patterns shared across demonstrations.

In addition, since IL is typically optimized by behavior cloning (BC), selected demonstrations should also provide stable and predictable action supervision. Atypical or noisy realizations of the same behavior pattern may introduce inconsistent actions under similar observations, making the conditional action prediction target harder to fit \cite{ross2010efficient,hussein2017imitation,belkhale2023data}. Therefore, effective data selection should not only decide \emph{which behavioral structures to preserve}, but also \emph{which concrete demonstrations to imitate}. This motivates an IL-friendly selection principle: after identifying useful behavior patterns, one should prefer stable central trajectory realizations that provide more consistent supervision for BC.

Based on these insights, we propose SIEVE, a structure-aware data selection framework for VLA imitation learning. SIEVE first discovers reusable atomic behavior primitives by segmenting trajectories at physically grounded interaction boundaries and clustering segment representations. Each trajectory is then represented as a primitive sequence, which defines a composition pattern and its adjacent transitions. SIEVE allocates the selection budget over composition-pattern buckets to maximize reusable primitive and transition exposure under diminishing returns, and then selects medoid trajectories within each bucket to obtain central, stable, and imitation-friendly realizations. This yields a compact subset that exposes reusable behavioral structure while providing predictable supervision for BC.
Our contributions are as follows:
\begin{itemize}
    \item We propose a primitive-compositional view of trajectory utility, realized by Primitive Discovery and Structural Exposure Allocation, which allocate selection budgets according to reuse-aware primitive and transition exposure under diminishing returns.
    \item We introduce Learning-Friendly Trajectory Selection, which selects medoid trajectories within each composition-pattern bucket to favor central, stable, and predictable realizations for behavior cloning.
    \item We present SIEVE, a structure-aware data selection method for VLA imitation learning, and demonstrate its effectiveness across datasets, benchmarks, and models. SIEVE can outperform full-data training using only 50\% of demonstrations and 50\% of training steps, while consistently improving over competitive baselines under multiple experimental settings.
\end{itemize}

%% file: sec/2_related.tex
\section{Related Work}
\subsection{Vision-Language-Action Models.} 
Recent advances in vision-language models (VLMs) \cite{shao2025large,bai2025qwen3,lin2025physbrain} have accelerated the development of Vision-Language-Action (VLA) models, establishing a scalable paradigm for language-conditioned robotic control \cite{zitkovich2023rt,o2024open,kim2024openvla}. Modern VLAs map visual observations and language instructions to robot actions through generative policy architectures, such as autoregressive action tokenization, diffusion, or flow-matching models \cite{zitkovich2023rt,chi2025diffusion,black2024pi_0,intelligence2025pi_,bjorck2025gr00t,kim2025fine,lian2026intentvla, intelligence2026pi}. These models are typically trained on large-scale demonstrations via imitation learning (IL), most commonly formulated as behavior cloning \cite{hussein2017imitation}.
However, robotic datasets often contain noisy, redundant, and unevenly distributed demonstrations, so scaling training data indiscriminately can waste computation and yield diminishing returns \cite{xing2025shortcut,sathyanarayan2025quality,belkhale2023data,lin2025data,o2024open}. This motivates selecting compact yet informative subsets for efficient VLA imitation learning.

\subsection{Data Selection for Imitation Learning.} 
Some related studies mainly focus on macroscopic dataset-level evaluation, mixing optimization across data sources and VLM data selection \cite{xiao2025data,belkhale2023data,hejna2024re,wu2026scalselect,zhou2026synthetic}.
Beyond dataset-level curation, existing IL data selection methods typically estimate sample utility either at the trajectory or state-action level. Trajectory-level methods use global signals such as representation similarity, reliability, downstream task feedback, or validation-loss-based influence estimation \cite{zhang2025scizor,chen2025curating,dass2025datamil,xu2026athena}, but they often collapse long-horizon demonstrations into a single score and, for feedback-based criteria, require costly additional model training.
State-action-level methods instead rely on local signals such as mutual information, task progress, or joint state-action similarity \cite{hejna2025robot,zhang2025scizor,yu2026frameskip}. However, these criteria are often either too myopic to capture coherent task semantics or primarily designed for local pruning.

%% file: sec/3_method.tex
\begin{figure*}[!thp]
  \centering
  \includegraphics[width=1\linewidth]{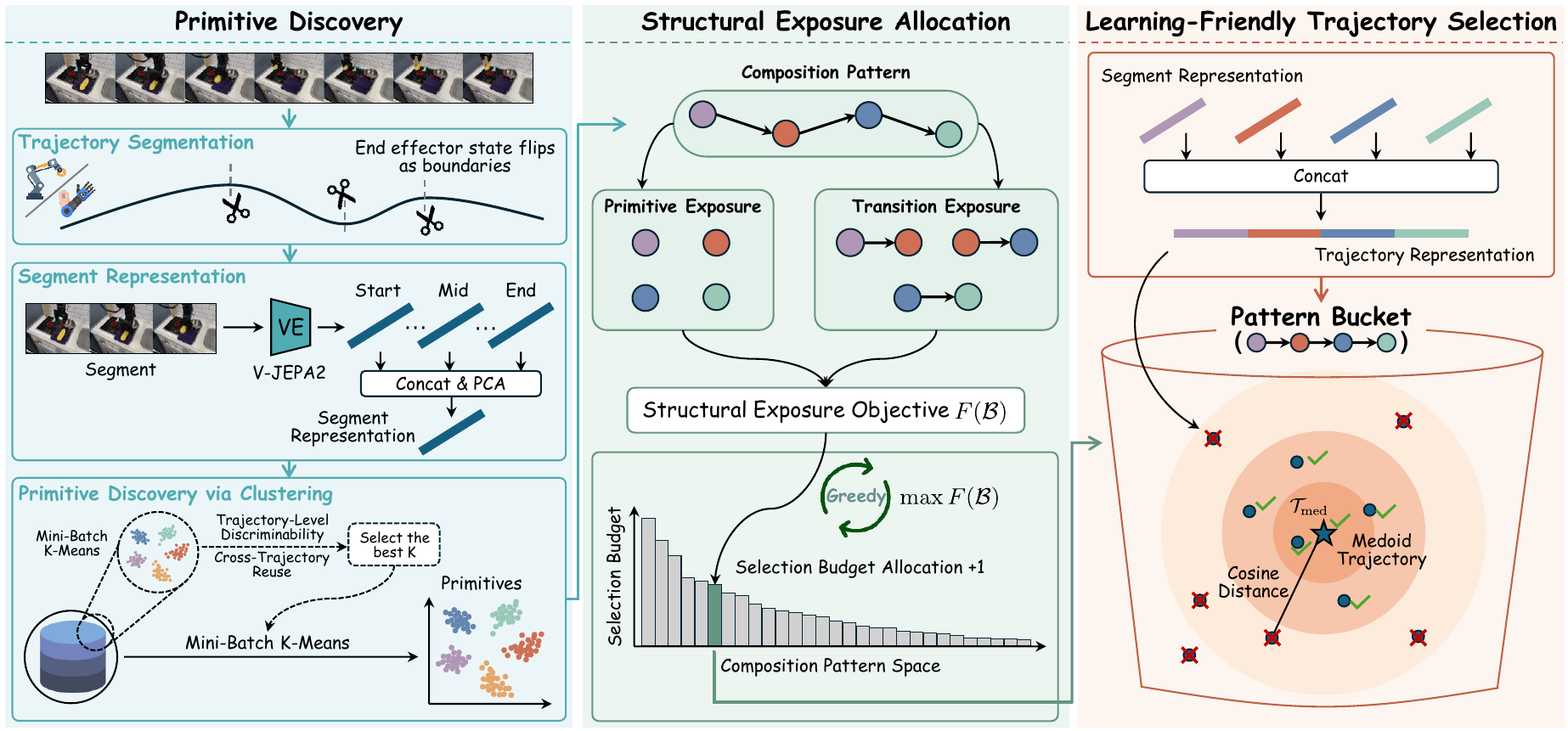}
   \caption{Overview of SIEVE. 1) Primitive Discovery: trajectories are segmented into primitives through representation learning and clustering. 2) Structural Exposure Allocation: a structural exposure objective estimates primitive composition coverage and allocates the selection budget across composition patterns. 3) Learning-Friendly Trajectory Selection: representative trajectories are selected within each composition-pattern bucket to construct a compact, diverse, and learning-friendly training subset.}
   \label{fig:overview}
\end{figure*}

\section{SIEVE}
We propose SIEVE, a structure-aware data selection method for imitation learning with VLA models. The key idea is to exploit the structural exposure of demonstrations for data selection. To this end, SIEVE first discovers visuo-motor primitives from trajectory segments and represents each trajectory as a composition of these primitives. It then allocates the selection budget based on structural exposure and selects representative, learning-friendly trajectories within each composition pattern. The resulting subset retains informative behavioral structures while improving the efficiency of imitation learning.

\subsection{Primitive Discovery}\label{sec:prim}

A demonstration trajectory is typically a composition of several reusable visuo-motor segments. This underlying structural prior motivates us to uncover a vocabulary of primitives from trajectories. We define a primitive as a reusable visuo-motor behavior unit discovered from trajectories, which serves as a proxy for a reusable behavioral subprogram.

\paragraph{Trajectory segmentation.}
Let $\mathcal{D}= \{\mathcal{T}_1,\dots,\mathcal{T}_N\}$ denote the original dataset, where each sample $\mathcal{T}_i$ is a demonstration trajectory.
For each trajectory $\mathcal{T}_i \in \mathcal{D}$, we segment it using end effector (gripper/dexterous-hand) state flips (i.e., grasp/release flips) as physically grounded interaction boundaries to obtain $L_i$ segments $\mathcal{T}_i=\{S_i^1,\dots,S_i^{L_i}\}$. To suppress spurious boundaries caused by transient actuation jitter, a state transition is accepted only if it persists for five consecutive frames.

\paragraph{Segment representation.}
We then extract the representation for each segment using a pretrained video encoder ($\operatorname{VE}$). Specifically, for each segment $S_i^j$ ($j\in \{1,\dots,L_i\}$), we uniformly sample 8 frames and encode them with V-JEPA2 \cite{assran2025v}.
As the start, middle, and end frames provide a compact summary of the segment's state evolution, we concatenate their representations to form a single feature that captures both semantic content and coarse temporal structure.
To suppress noise, improve clustering stability, and reduce computational cost, we further reduce the representation to 256 dimensions with PCA:
\begin{equation}
\begin{aligned}
    z_i^j \leftarrow &\operatorname{PCA}(\operatorname{Concat}(\operatorname{VE}(S_i^j)_\texttt{start};\\&\operatorname{VE}(S_i^j)_\texttt{mid};\operatorname{VE}(S_i^j)_\texttt{end})).
\end{aligned}
\end{equation}

\paragraph{Primitive discovery via clustering.}
We discover primitives by clustering the segment representations using MiniBatch K-Means \cite{sculley2010web}. Rather than manually specifying the number of clusters, we select $K$ automatically on a randomly sampled subset of trajectories. We seek a primitive vocabulary that is both \emph{reusable} across trajectories and \emph{discriminative} at the trajectory level: useful primitives should recur across demonstrations, while not collapsing structurally different trajectories into nearly identical primitive composition patterns.

For each candidate $K$, we first cluster the segment representations from the randomly sampled subset and summarize each trajectory $\mathcal{T}_i$ by the set of clusters it covers:
\begin{equation}
    \mathcal{C}_i=\{c_i^1,\dots,c_i^{|\mathcal{C}_i|}\} \subseteq \{1,\dots,K\},
\end{equation}
where each $c_i^u \in \{1,\dots,K\}$ is a cluster index covered by at least one segment in $\mathcal{T}_i$.
Based on these trajectory-level cluster sets, we evaluate each candidate vocabulary using a reuse-aware criterion comprising two components. The first, \emph{trajectory-level discriminability} $\mathcal{J}$, measures how distinct different trajectories remain after being represented by cluster coverage.
We compute $\mathcal{J}$ as the median of the average pairwise Jaccard similarities over the cluster sets $\mathcal{C}_i$ across all sampled trajectories:
\begin{equation}
    \mathcal{J} = \operatorname{median}_{i=1}^{n}\, \frac{1}{n-1}\sum_{j \ne i}^n \operatorname{Jaccard}(\mathcal{C}_i,\mathcal{C}_j).
\end{equation}
A lower $\mathcal{J}$ indicates better preservation of structural distinctions between trajectories.

The second component, \emph{cross-trajectory reuse} $\mathcal{R}$, quantifies how broadly each discovered cluster recurs across trajectories.
$\mathcal{R}$ is defined as the median occurrence count of each cluster across all trajectories:
\begin{equation}
    \mathcal{R} = \operatorname{median}_{k=1}^{K}\, \sum_{i=1}^{n} \mathbf{1}[\,k \in \mathcal{C}_i\,].
\end{equation}
A higher $\mathcal{R}$ implies that the discovered primitives capture widely applicable behavior units.

We choose the number of clusters by maximizing
\begin{equation}
    K^* = \arg\max_{K}
    \left[
    \bigl(1 - \mathcal{J}\bigr)
    \log \mathcal{R}
    \right].
\end{equation}
This criterion favors primitive vocabularies that are broadly reused across trajectories while still preserving trajectory-level discriminability.
After selecting $K^*$, we rerun MiniBatch K-Means on all segment representations, and each resulting cluster is treated as a discovered primitive.

\subsection{Structural Exposure Allocation}

After primitive discovery, each trajectory $\mathcal{T}_i$ is represented by an ordered primitive sequence, referred to as a \emph{composition pattern}:
\begin{equation}
    P_i = [c_i^1, \dots, c_i^{|P_i|}],
\end{equation}
where each $c_i^m \in \mathcal{C}_i$ is a discovered primitive.

We further define a \emph{transition} as the local compositional interface between two adjacent primitives in a composition pattern. For trajectories with $|P_i|\ge2$, the transitions are
\begin{equation}
    e_i^j=(c_i^j\rightarrow c_i^{j+1}),
    \qquad
    j=1,\dots,|P_i|-1.
\end{equation}
For trajectories consisting of a single primitive, we introduce a terminal null state $\varnothing$ and define the terminal transition:
\begin{equation}
    e_i=(c_i^1\rightarrow\varnothing),
\end{equation}
so that every trajectory contains at least one transition.

Let $\mathcal{P}=\{P^{(1)},\dots,P^{(|\mathcal{P}|)}\}$ denote the set of unique composition patterns in the dataset, where $P^{(\ell)}$ denotes the $\ell$-th unique composition pattern, corresponding to a pattern bucket containing trajectories with the same composition pattern.
This composition-pattern space provides the structural reference for measuring how broadly reusable structures are shared across demonstrations. In SIEVE, we capture such reusable structures through primitive composition, where trajectories expose both atomic behavior primitives and the transition interfaces connecting them.
To effectively learn reusable structures, the model should be repeatedly exposed not only to primitives themselves but also to how they are composed through transitions. In our setting, these transition interfaces often coincide with critical state changes (e.g., gripper grasp/release) in the task, and therefore provide informative cues for local behavior progression. Moreover, primitives and transitions that are reused across more composition patterns support a broader range of executable behaviors and should be preferentially preserved.

To retain diverse and important structural information, we allocate the selection budget over the composition-pattern space. Let
\begin{equation}
    \mathcal{B}=[b_1,\dots,b_{|\mathcal{P}|}]
\end{equation}
denote the allocation vector, where $b_\ell$ is the number of trajectories retained for composition pattern $P^{(\ell)}$.
We optimize the budget allocation $\mathcal{B}$ by maximizing the following structural exposure objective:
\begin{equation}
    F(\mathcal{B})
    =
    \sum_{c \in \mathcal{C}} w_c \log\bigl(1+n_c(\mathcal{B})\bigr)
    +
    \sum_{e \in \mathcal{E}} w_e \log\bigl(1+n_e(\mathcal{B})\bigr),
\end{equation}
where $\mathcal{C}$ and $\mathcal{E}$ denote the sets of all discovered primitives and transitions, respectively. Here, $n_c(\mathcal{B})$ and $n_e(\mathcal{B})$ denote the numbers of occurrences of primitive $c$ and transition $e$ among the selected trajectories under allocation $\mathcal{B}$.
The primitive and transition weights are defined by their reuse frequency over the composition-pattern space:
\begin{equation}
\begin{aligned}
    w_c &= \frac{q_c}{|\mathcal{P}|}, \qquad
    q_c = \left| \{\, P \in \mathcal{P} : c \in P \,\} \right|, \\
    w_e &= \frac{q_e}{|\mathcal{P}|}, \qquad
    q_e = \left| \{\, P \in \mathcal{P} : e \in P \,\} \right|.
\end{aligned}
\end{equation}
Here, $q_c$ and $q_e$ denote the numbers of composition patterns containing primitive $c$ and transition $e$, respectively. Consequently, primitives and transitions that participate in more composition patterns receive larger weights. The logarithmic utility introduces diminishing returns, encouraging the budget to expose the model to a broader set of reusable structures rather than repeatedly reinforcing the same ones.

We optimize $F(\mathcal{B})$ greedily. Starting from $\mathcal{B}^{(0)}=\mathbf{0}$, each iteration allocates one additional sample to the composition pattern with the largest marginal gain:
\begin{equation}
    \Delta(P^{(\ell)}\mid\mathcal{B})
    =
    F(\mathcal{B}+\mathbf{b}_\ell)-F(\mathcal{B}),
\end{equation}
where $\mathbf{b}_\ell$ is the one-hot allocation vector that increases the budget of pattern $P^{(\ell)}$ by one. We select
\begin{equation}
\ell^\star
=
\mathop{\mathrm{arg\,max}}_{\ell\in\{1,\dots,|\mathcal{P}|\}}
\Delta(P^{(\ell)}\mid\mathcal{B}),
\end{equation}
and update
\begin{equation}
    \mathcal{B}
    \leftarrow
    \mathcal{B}
    +
    \mathbf{b}_{\ell^\star}.
\end{equation}
After the budget is exhausted, $\mathcal{B}$ specifies how many trajectories should be retained from each pattern bucket. The actual trajectories are then selected within each pattern bucket in the next stage.

\subsection{Learning-Friendly Trajectory Selection}
Given the pattern-level budget allocation, this stage selects representative and learning-friendly trajectories within each pattern bucket.
Behavior cloning trains the policy by minimizing
\begin{equation}
\mathcal{L}_{\mathrm{BC}}(\theta)
=
\mathbb{E}_{(s,a)\sim\mathcal D}
[-\log \pi_\theta(a|s)],
\end{equation}
where $(s,a)$ is a state-action pair sampled from the demonstration dataset $\mathcal{D}$, and $\pi_\theta(a|s)$ is the probability assigned by the policy $\pi_\theta$ to action $a$ under state $s$.
Demonstrations with more consistent state-action mappings provide clearer supervision and are easier for imitation learning to optimize. Since directly estimating conditional action entropy is impractical, we use representation-space centrality as a practical proxy: trajectories near the center of a composition pattern are less likely to be outliers or ambiguous demonstrations, and thus tend to provide more stable supervision.

For each trajectory $\mathcal{T}_i$ with composition pattern $P^{(\ell)}$, we construct a trajectory representation $x_i$ by concatenating the segment representations along its primitive sequence:
\begin{equation}
    x_i = \operatorname{Concat}(z_i^1; \dots; z_i^{|P_i|}).
\end{equation}
Within each composition pattern, we compare trajectories using cosine similarity and identify the medoid trajectory $\mathcal{T}_{\mathrm{med}}$, which has the largest aggregate similarity $S_i=\sum_{j\neq i}\cos(x_i,x_j)$ to other trajectories in the same pattern. We then rank trajectories by their distance to the medoid, defined as $d_i=1-\cos(x_i,x_{\mathrm{med}})$, and retain the $\mathcal{B}[\ell]$ trajectories with the smallest distances, where $\mathcal{B}[\ell]$ is the budget assigned to composition pattern $P^{(\ell)}$ by Structural Exposure Allocation. This selects trajectories closest to the pattern bucket center, yielding representative and learning-friendly demonstrations.

%% file: sec/4_exp.tex
\begin{table*}[!thp]
\centering
\renewcommand{\arraystretch}{1}
\setlength\tabcolsep{8pt} 
\resizebox{\linewidth}{!}
{
\begin{tabular}{l|c|cccc|c}
\toprule
\textbf{Method} & \textbf{Training Steps} &\textbf{Stack Green Cube On Yellow Cube} & \textbf{Put Carrot On Plate} & \textbf{Put Spoon On Table Cloth} & \textbf{Put Eggplant In Basket} & \textbf{Avg.}\\
\midrule
Full-Training & 50K & 22.9 & 53.1 & 68.8 & 62.5 & 51.8 \\
\midrule
\multicolumn{7}{c}{\textit{\textbf{Selection Budget: 26.5K (50\%)}}} \\
\midrule
Random & \multirow{4}{*}{25K (50\%)} & 20.8 & 41.7 & 64.6 & 31.3 & 39.6 \\
DemInf &  & 11.5 & 37.5 & 58.3 & 65.6 & 43.2 \\
SCIZOR &  & 13.5 & 36.5 & 68.8 & 90.6 & 52.2 \\
\rowcolor{navyblue!10}SIEVE (Ours) &  & 25.0 & 54.2 & 70.8 & 75.0 & 56.3 \\
\midrule
Random & \multirow{4}{*}{50K (100\%)} & 25.0 & 36.5 & 66.7 & 33.3 & 40.4 \\
DemInf &  & 12.5 & 40.6 & 69.8 & 63.5 & 46.6 \\
SCIZOR &  & 16.7 & 39.6 & 72.9 & \textbf{92.7} & 55.5 \\
\rowcolor{navyblue!10}SIEVE (Ours) &  & \textbf{29.2} & 57.3 & 75.0 & 76.0 & 59.4 \\
\midrule
\multicolumn{7}{c}{\textit{\textbf{Selection Budget: 37.1K (70\%)}}} \\
\midrule
Random & \multirow{4}{*}{35K (70\%)} & 16.7 & 50.0 & 74.0 & 37.5 & 44.6 \\
DemInf &  & 18.8 & 51.0 & 78.1 & 72.9 & 55.2 \\
SCIZOR &  & 20.8 & 41.7 & 72.9 & 91.7 & 56.8 \\
\rowcolor{navyblue!10}SIEVE (Ours) &  & 22.9 & 57.3 & 81.3 & 87.5 & 62.3 \\
\midrule
Random & \multirow{4}{*}{50K (100\%)} & 12.5 & 51.0 & 77.1 & 46.9 & 46.9 \\
DemInf &  & 17.7 & 57.3 & 79.2 & 74.0 & 57.1 \\
SCIZOR &  & 19.8 & 45.8 & 76.0 & 90.6 & 58.1 \\
\rowcolor{navyblue!10}SIEVE (Ours) &  & 21.9 & \textbf{58.3} & \textbf{86.5} & 83.3 & \textbf{62.5} \\
\bottomrule
\end{tabular}
}
\caption{Performance comparison with baselines on SimplerEnv-WidowX using Qwen3-VL-4B-GR00T. We compare SIEVE with baselines under 50\% and 70\% selection budgets, at different training steps. ``Avg.'' denotes the average success rate.}
\label{tab:exp_main}
\end{table*}

\section{Experiments}

\subsection{Experiment Setup}
\textbf{Datasets and Evaluation.}
Here we evaluate SIEVE on three representative robot imitation learning datasets to assess both its data selection effectiveness and its applicability across different embodiments and environments. Unless otherwise specified, Bridge-V2 with SimplerEnv-WidowX serves as the default training and evaluation setting.

\begin{itemize}
    \item \emph{Bridge-V2}: We train models on the Bridge-V2 dataset \cite{walke2023bridgedata}, a real-world subset of Open X-Embodiment, containing approximately 53K demonstration trajectories collected with a WidowX robot equipped with a parallel gripper. Policies are evaluated in SimplerEnv-WidowX \cite{li2024evaluating}, which includes four manipulation tasks: \emph{Stack Cube}, \emph{Put Carrot}, \emph{Put Spoon}, and \emph{Put Eggplant}. Evaluation is conducted under multiple unseen kitchen backgrounds and randomized object configurations, providing a challenging out-of-distribution (OOD) benchmark.
    \item \emph{Fractal}: We further evaluate SIEVE on the Fractal subset of Open X-Embodiment \cite{o2024open}, which contains approximately 87K real-world manipulation trajectories collected with a Google Robot manipulator. Policies are evaluated in SimplerEnv-GoogleRobot \cite{li2024evaluating} on three manipulation tasks: \emph{Grasp Coke Can}, \emph{Move Near}, and \emph{Close/Open Drawer}.
    \item \emph{GR00T-X-Sim}: We additionally evaluate SIEVE on the downsampled Humanoid Robot Tabletop Manipulation subset of GR00T-X-Embodiment-Sim \cite{bjorck2025gr00t}, which contains 24K simulated demonstration trajectories collected with a humanoid robot equipped with dexterous hands. Policies are evaluated in RoboCasa-GR1 \cite{nasiriany2024robocasa}, a benchmark comprising 24 tabletop manipulation tasks across diverse scene layouts and object configurations.
\end{itemize}

\textbf{Models.} We evaluate SIEVE on two representative VLA models, Qwen3-VL-4B-GR00T and Qwen3-VL-4B-OFT. Unless otherwise specified, all main experiments and ablation studies are conducted using Qwen3-VL-4B-GR00T.

\begin{itemize}
    \item \emph{Qwen3-VL-4B-GR00T} \cite{bjorck2025gr00t}: This model combines the Qwen3-VL-4B vision-language backbone \cite{bai2025qwen3} with a GR00T-style flow-matching policy head that predicts continuous robot actions through conditional flow matching. 
    \item \emph{Qwen3-VL-4B-OFT} \cite{kim2025fine}: This model uses the Qwen3-VL-4B backbone and adopts the OpenVLA-OFT action decoding recipe, which performs parallel continuous action prediction with an MLP-based prediction head trained using an L1 regression objective.
\end{itemize}

\textbf{Baselines.}
We compare SIEVE with the following baselines. All experiments use the same training hyperparameters for a fair comparison (see Appendix for details):
\begin{itemize}
\item \emph{Full-Training}: trains on the complete dataset without data selection and serves as the full-data reference.
\item \emph{Random}: uniformly samples demonstrations from the original training set.
\item \emph{DemInf}~\cite{hejna2025robot}: selects demonstrations according to state-action mutual information estimates.
\item \emph{SCIZOR}~\cite{zhang2025scizor}: filters low-quality data by identifying redundant trajectories and suboptimal state-action pairs.
\end{itemize}

\subsection{Main Results}

Table~\ref{tab:exp_main} summarizes the main results on Bridge-V2 using Qwen3-VL-4B-GR00T. We evaluate two selection budgets, 50\% and 70\%, corresponding to 26.5K and 37.1K demonstrations, respectively. For each budget, we consider two training schedules. The first scales the number of training steps proportionally to the selected data size (25K and 35K steps for the 50\% and 70\% budgets, respectively), reducing both training data and computation. The second trains each selected subset for the same 50K training steps as Full-Training. Comparing these two settings allows us to disentangle improvements brought by data selection from those potentially arising from increased optimization per sample under a fixed training budget.

As shown in Table~\ref{tab:exp_main}, SIEVE consistently achieves the highest average success rate across all selection budgets and training schedules. Notably, using only 50\% of the demonstrations and 25K training steps, SIEVE attains an average success rate of 56.3\%, outperforming Full-Training (51.8\%), which uses the complete dataset and twice the training steps. This result indicates that SIEVE is able to identify a compact subset that is more effective for imitation learning while substantially reducing both training data and computation. 

Compared with existing data selection baselines, SIEVE consistently delivers the best performance. Under the 50\% selection budget, SIEVE achieves average success rates of 56.3\% and 59.4\% under the proportional and 50K training schedules, outperforming SCIZOR (52.2\% and 55.5\%, respectively). Under the 70\% budget, SIEVE further achieves average success rates of 62.3\% and 62.5\% with 35K and 50K training steps, respectively, again outperforming all competing methods.
SIEVE also exhibits consistently stronger per-task performance. It outperforms Random on every task under all evaluation settings, suggesting that the performance gains arise from selecting structurally informative demonstrations rather than simply reducing the training set size. Compared with SCIZOR, SIEVE produces more balanced performance across tasks. While SCIZOR performs particularly well on \emph{Put Eggplant In Basket}, its performance drops noticeably on \emph{Stack Green Cube On Yellow Cube} and \emph{Put Carrot On Plate}. In contrast, SIEVE maintains competitive performance across all four tasks, resulting in the best overall average success rate.

Since SimplerEnv evaluates policies under substantial visual and environmental variations relative to Bridge-V2, these results further demonstrate that the subsets selected by SIEVE generalize well under distribution shift.

\subsection{Extended Results}

\begin{table}[!thp]
\centering
\renewcommand{\arraystretch}{1}
\setlength\tabcolsep{14pt} 
\resizebox{\linewidth}{!}
{
\begin{tabular}{l|c|c|c}
\toprule
\textbf{Method} & \textbf{Bridge-V2} & \textbf{Fractal} & \textbf{GR00T-X-Sim}\\
\midrule
Full-Training & 51.8 & 75.0 & 52.7  \\
\midrule
Random & 39.6 & 55.6 & 53.5  \\
DemInf& 43.2 & 67.4 & 53.8  \\
SCIZOR & 52.2 & 71.9 & 54.2 \\
\rowcolor{navyblue!10}SIEVE (Ours) & 56.3 & 76.4 & 54.8 \\
\bottomrule
\end{tabular}
}
\caption{Performance across different datasets using Qwen3-VL-4B-GR00T. Full-Training uses the complete dataset, while all selection methods use 50\% of the data and 50\% of the training steps. We report average success rates (\%).}
\label{tab:exp_diff_datasets}
\end{table}

\textbf{Performance Across Different Datasets.}
We further evaluate SIEVE across different training datasets and evaluation benchmarks. For Bridge-V2, Full-Training is trained for 50K steps, whereas for Fractal and GR00T-X-Sim, Full-Training is trained for 100K steps. For all data selection methods, we retain 50\% of the training data and train for 50\% of the corresponding Full-Training steps (25K for Bridge-V2 and 50K for Fractal and GR00T-X-Sim).
As shown in Table~\ref{tab:exp_diff_datasets} (see the Appendix for detailed results), SIEVE achieves the highest average success rate across all three settings. On Bridge-V2 and Fractal, SIEVE achieves average success rates of 56.3\% and 76.4\%, outperforming Full-Training (51.8\% and 75.0\%, respectively), while using only half of the training data and half of the training steps. On GR00T-X-Sim, Random selection already slightly outperforms Full-Training. One possible explanation is that both GR00T-X-Sim and RoboCasa-GR1 are simulated tabletop manipulation environments, making RoboCasa-GR1 relatively in-domain for GR00T-X-Sim. In this setting, training on a smaller subset may provide more effective optimization per demonstration. Nevertheless, SIEVE still achieves the highest average success rate among all methods.
These results demonstrate that SIEVE generalizes effectively across different datasets, robot embodiments, and evaluation benchmarks.

\begin{table}[!tbhp]
\centering
\renewcommand{\arraystretch}{1.2}
\setlength\tabcolsep{4pt} 
\resizebox{\linewidth}{!}
{
\begin{tabular}{l|cccc|c}
\toprule
\textbf{Method}  & \makecell{\textbf{Stack Green Cube} \\ \textbf{On Yellow Cube}} & 
\makecell{\textbf{Put Carrot} \\ \textbf{On Plate}} & 
\makecell{\textbf{Put Spoon} \\ \textbf{On Table Cloth}} & 
\makecell{\textbf{Put Eggplant} \\ \textbf{In Basket}} & \textbf{Avg.}\\
\midrule
\multicolumn{6}{c}{\textit{\textbf{Qwen3-VL-4B-GR00T}}} \\
\midrule
 Full-Training &  22.9 & 53.1 & 68.8 & 62.5 & 51.8 \\
Random &  20.8 & 41.7 & 64.6 & 31.3 & 39.6 \\
\rowcolor{navyblue!9}SIEVE (Ours)  & 25.0 & 54.2 & 70.8 & 75.0 & 56.3 \\
\midrule
\multicolumn{6}{c}{\textit{\textbf{Qwen3-VL-4B-OFT}}} \\
\midrule
 Full-Training &  21.9 & 25.0 & 45.8 & 62.5 & 38.8 \\
Random &  9.4 & 33.3 & 16.6 & 45.8 & 26.3 \\
\rowcolor{navyblue!9}SIEVE (Ours)  & 21.9 & 58.3 & 50.0 & 95.8 & 56.5 \\
\bottomrule
\end{tabular}
}
\caption{Performance across different models on SimplerEnv-WidowX. Full-Training uses the complete dataset with 50K training steps, while all selection methods use 50\% of the data and 50\% of the training steps.}
\label{tab:exp_diff_models}
\end{table}

\textbf{Performance Across Different Models.}
We further evaluate whether SIEVE-selected data remains effective across different VLA models. Since this experiment aims to evaluate model generalization rather than compare all data selection methods, we compare SIEVE with Full-Training and Random under the same 50\% selection budget.
As shown in Table~\ref{tab:exp_diff_models}, SIEVE consistently achieves the highest average success rate on both Qwen3-VL-4B-GR00T and Qwen3-VL-4B-OFT. On Qwen3-VL-4B-GR00T, SIEVE achieves an average success rate of 56.3\%, outperforming both Random (39.6\%) and Full-Training (51.8\%). On Qwen3-VL-4B-OFT, SIEVE further achieves an average success rate of 56.5\%, outperforming Random (26.3\%) and Full-Training (38.8\%). The gains are particularly evident on tasks such as \emph{Put Carrot On Plate} and \emph{Put Eggplant In Basket}, where SIEVE-selected data consistently achieves substantially higher success rates than both reference methods. These results indicate that the effectiveness of SIEVE is not tied to a particular VLA model.

\begin{table}[!thp]
\centering
\renewcommand{\arraystretch}{1.2}
\setlength\tabcolsep{4pt} 
\resizebox{\linewidth}{!}
{
\begin{tabular}{l|cccc|c}
\toprule
\textbf{Method}  &\makecell{\textbf{Stack Green Cube} \\ \textbf{On Yellow Cube}} & 
\makecell{\textbf{Put Carrot} \\ \textbf{On Plate}} & 
\makecell{\textbf{Put Spoon} \\ \textbf{On Table Cloth}} & 
\makecell{\textbf{Put Eggplant} \\ \textbf{In Basket}} & \textbf{Avg.}\\
\midrule
\multicolumn{6}{c}{\textit{\textbf{Ablation of Structural Exposure Allocation}}} \\
\midrule
 w/o Trans.  & 22.9 & 47.9 & 67.7 & 64.6 & 50.8 \\
 w/o Prim. & 25.0 & 51.0 & 68.8 & 61.5 & 51.6 \\
\midrule
\multicolumn{6}{c}{\textit{\textbf{Ablation of Learning-Friendly Trajectory Selection}}} \\
\midrule
Most-Dissim  & 18.8 & 45.8 & 64.6 & 31.3 & 40.1 \\
Random  & 29.2 & 36.5 & 77.1 & 71.9 & 53.7 \\
\midrule
\rowcolor{navyblue!10}SIEVE (Ours) & 25.0 & 54.2 & 70.8 & 75.0 & 56.3 \\
\bottomrule
\end{tabular}
}
\caption{
Ablation study on SimplerEnv-WidowX using Qwen3-VL-4B-GR00T under the 50\% selection budget and 50\% of the training steps.
``w/o Trans.'' and ``w/o Prim.'' remove the transition and primitive exposure terms, respectively.
``Most-Dissim'' selects the most dissimilar samples to maximize within-bucket trajectory diversity, while ``Random'' uses random within-bucket selection.
}
\label{tab:exp_ablation}
\end{table}

\subsection{Ablation Study and Further Analysis}

\textbf{Ablation of Structural Exposure Allocation.}
We ablate the two structural exposure terms in SIEVE. Removing the transition or primitive exposure term decreases the average success rate from 56.3\% to 50.8\% and 51.6\%, respectively, indicating that both primitives and transitions contribute to effective data selection. The larger performance drop without transition exposure further suggests that transition interfaces provide important supervision for modeling how reusable behavior units are connected during task execution.

\textbf{Ablation of Learning-Friendly Trajectory Selection.}
We ablate the within-bucket sample selection strategy. Replacing the proposed selection with Most-Dissim, which selects trajectories with the lowest aggregate cosine similarity to other samples to maximize trajectory diversity, reduces the average success rate from 56.3\% to 40.1\%, indicating that atypical trajectories within the same composition pattern are less suitable for imitation learning. Random within-bucket selection performs better than most-dissimilar selection but still trails SIEVE, improving the average success rate only from 53.7\% to 56.3\%. These results validate our design choice of selecting representative and stable demonstrations within each composition-pattern bucket.


\begin{figure}[!th]
  \centering
  \includegraphics[width=1\linewidth]{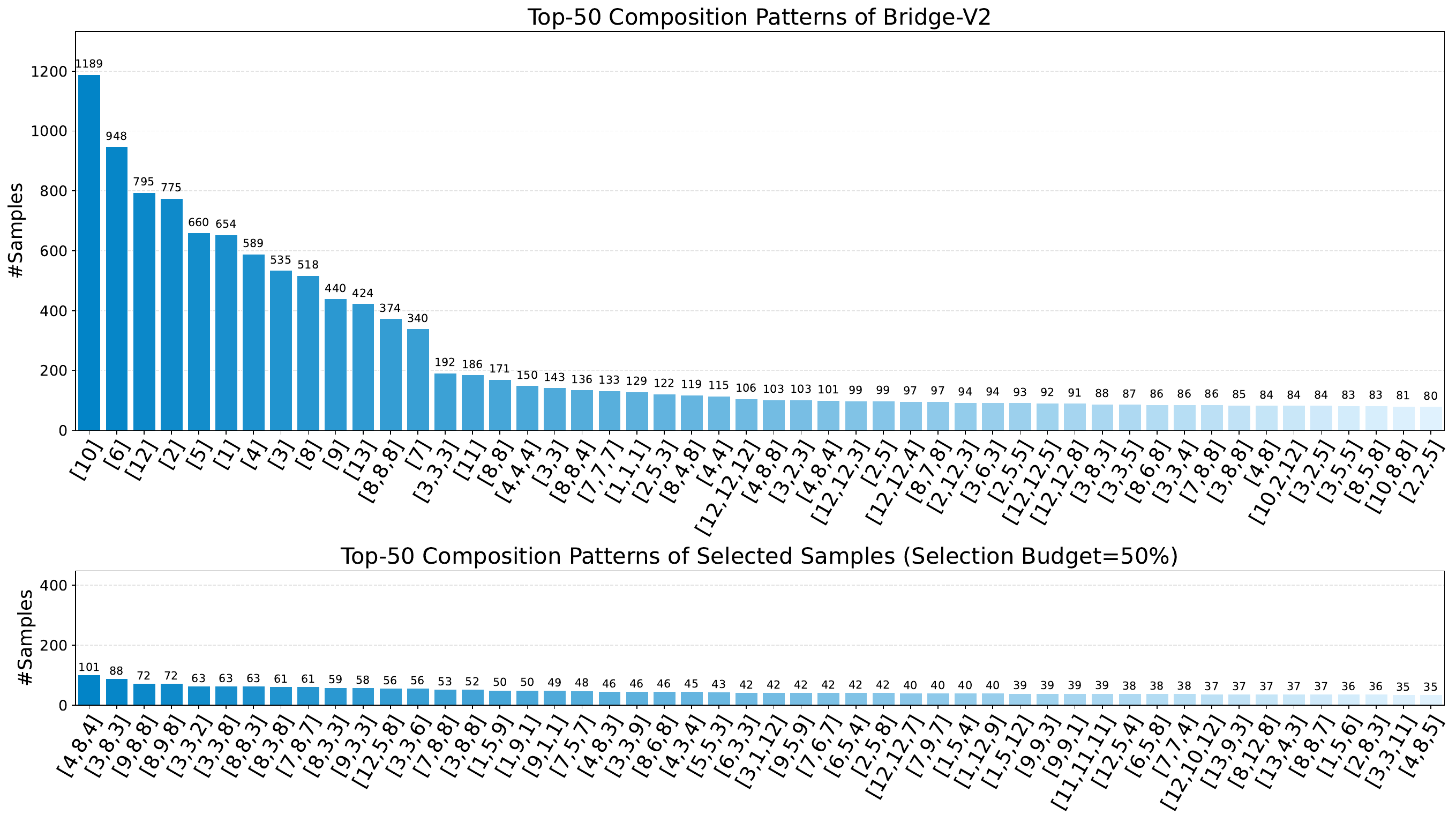}
   \caption{
Distribution of top-50 composition patterns before and after SIEVE selection (Selection budget=50\%) on Bridge-V2.
Each x-axis label denotes a composition pattern, represented as an ordered list of primitive IDs.
}
   \label{fig:distribution}
\end{figure}

\textbf{Composition Pattern Redistribution.}
Figure~\ref{fig:distribution} illustrates how SIEVE reshapes the composition-pattern distribution on Bridge-V2. The original dataset is dominated by a small number of high-frequency patterns, many of which consist of a single primitive, indicating limited compositional diversity. After selection, the distribution becomes substantially more balanced, and the most frequent selected patterns are predominantly multi-primitive sequences, covering a wider range of primitives and transitions. This redistribution suggests that SIEVE shifts the selected subset away from frequency-dominated patterns toward structurally richer composition patterns, thereby exposing more reusable primitives and transitions for imitation learning.

%% file: sec/5_conclusion.tex
\section{Conclusion}

We introduced SIEVE, a structure-aware data selection method for VLA imitation learning. SIEVE selects demonstrations by exposing reusable primitive compositions and transition interfaces, while favoring central and stable realizations within each composition pattern for behavior cloning. Across multiple datasets, benchmarks, and VLA models, SIEVE consistently improves over competitive baselines and can outperform full-data training using fewer demonstrations and training steps. These results highlight the importance of selecting data according to reusable structure, offering a practical route toward more efficient VLA imitation learning.

%% file: sec/6_appendix.tex
\section{Appendix}
\subsection{Training Hyperparameters}
All experiments are conducted using the same training hyperparameters summarized in Table~\ref{tab:app_hyper}, ensuring a fair comparison across different methods.
All models are trained on 8 NVIDIA H100 (80GB) GPUs.

\begin{table}[!thp]
\centering
\renewcommand{\arraystretch}{1.2}
\setlength\tabcolsep{14pt} 
\resizebox{0.7\linewidth}{!}
{
\begin{tabular}{l|c}
\toprule
\textbf{Hyperparameter} & \textbf{Value}\\
\midrule
Optimizer & AdamW \\
Learning rate (VLM) & 1e-5 \\
Learning rate (Action Head) & 1e-4 \\
LR scheduler & Cosine decay \\
Warmup ratio & 10\% \\
AdamW $(\beta_1,\beta_2)$ & $(0.9,\,0.95)$ \\
Weight decay & 1e-8 \\
Gradient clipping & 1.0 \\
Per-device batch size & 16 \\
Gradient accumulation & 1 \\
Mixed precision & BF16 \\
Distributed training & DeepSpeed ZeRO-2 \\
\bottomrule
\end{tabular}
}
\caption{Training hyperparameters.}
\label{tab:app_hyper}
\end{table}

\subsection{Implementation Details of Primitive Discovery via Clustering.}
\begin{figure}[!t]
  \centering
  \captionsetup[subfigure]{skip=-1pt} 
  \begin{subfigure}{\linewidth}
    \centering
    \includegraphics[width=0.8\linewidth]{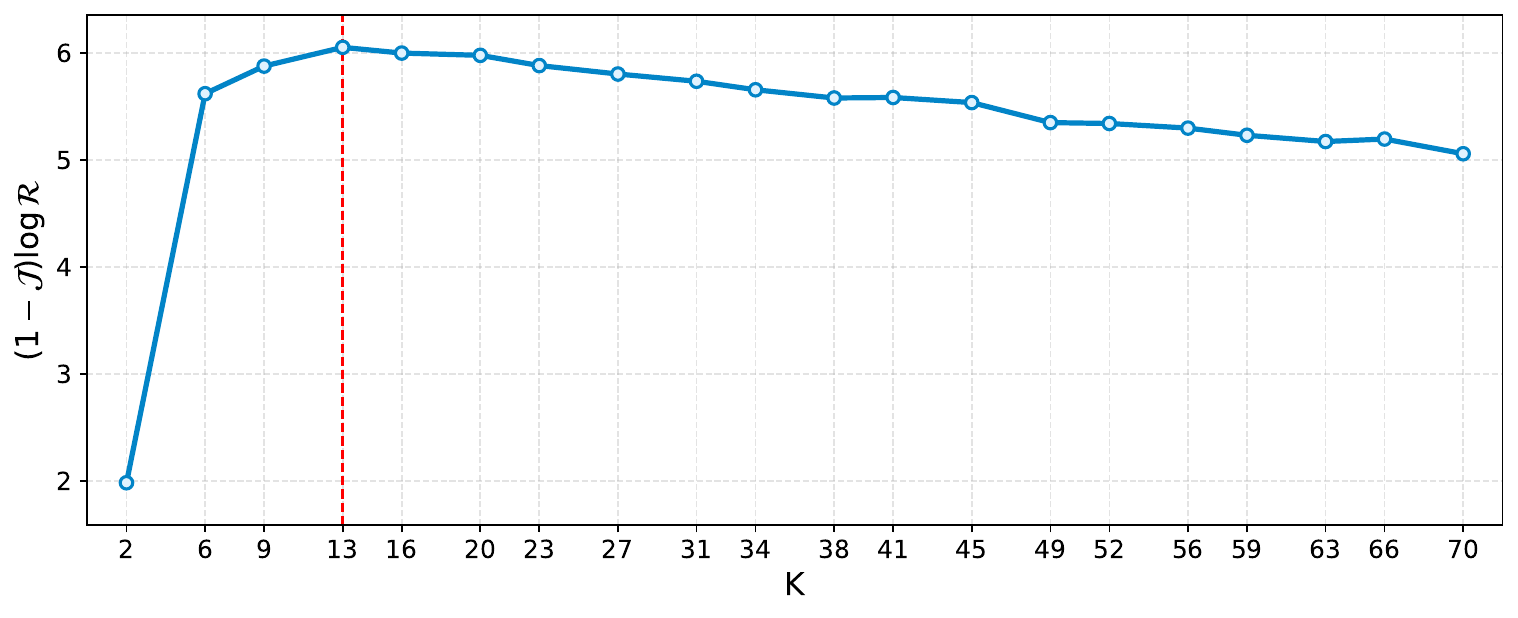}
    \caption{Bridge-V2}
    \label{fig:app_k_bridge_a}
  \end{subfigure}
  
  \vspace{0pt} 
  
  \begin{subfigure}{\linewidth}
    \centering
    \includegraphics[width=0.8\linewidth]{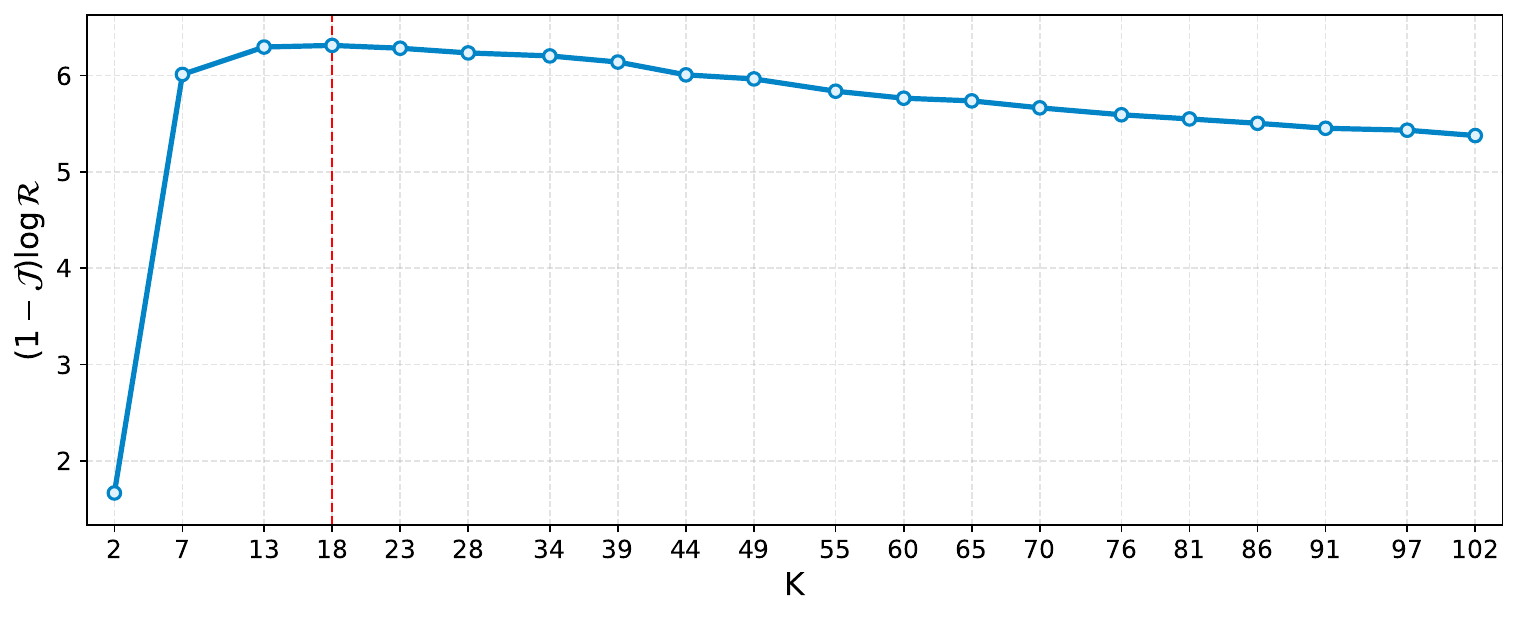} 
    \caption{Fractal}
    \label{fig:app_k_bridge_b}
  \end{subfigure}
  
  \vspace{0pt} 
  
  \begin{subfigure}{\linewidth}
    \centering
    \includegraphics[width=0.8\linewidth]{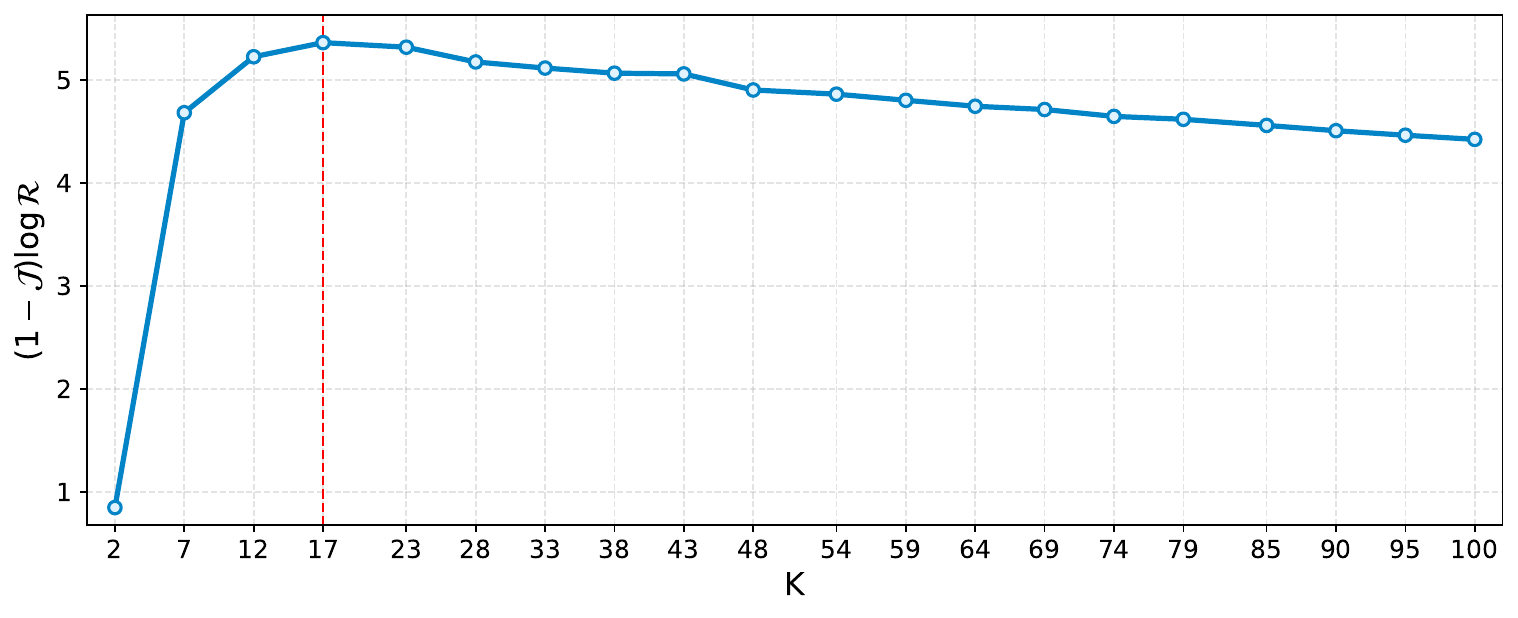} 
    \caption{GR00T-X-Sim}
    \label{fig:app_k_bridge_c}
  \end{subfigure}
  
  \vspace{-5pt} 
  \caption{Selection of the number of primitive clusters $K$ on different datasets. Following the practical protocol used in our experiments, we randomly sample 10\% of each training dataset and uniformly evaluate 20 candidate $K$ values within a reasonable search range. The optimal $K$ is selected by maximizing the clustering score $(1-\mathcal{J})\log\mathcal{R}$, and is indicated by the red dashed line.}
  \label{fig:app_k}
\end{figure}

To determine the number of primitive clusters, we adopt a practical search strategy. Specifically, for each dataset, we randomly sample 10\% of the demonstrations and uniformly evaluate 20 candidate values of $K$ within a reasonable search range. For each candidate, we perform primitive clustering and compute the clustering score $(1-\mathcal{J})\log\mathcal{R}$ defined in the \emph{Primitive discovery via clustering} section. The value of $K$ that maximizes this score is selected for all subsequent experiments.
Figure~\ref{fig:app_k} shows the search results on Bridge-V2, Fractal, and GR00T-X-Sim. The clustering score exhibits a clear peak on all three datasets, suggesting that the proposed criterion provides a stable and practical heuristic for determining the number of primitive clusters.

\subsection{Details on Performance Across Different Datasets}\label{sec:app_datasets}

Tables~\ref{tab:app_fractal} and~\ref{tab:app_gr00t} present the detailed results corresponding to Table~\ref{tab:exp_diff_datasets} in the main paper. Consistent with the main results, SIEVE achieves the highest average success rate on both Fractal and GR00T-X-Sim while maintaining competitive performance across individual evaluation categories.

\begin{table}[!tb]
\centering
\renewcommand{\arraystretch}{1.2}
\setlength\tabcolsep{4pt} 
\resizebox{\linewidth}{!}
{
\begin{tabular}{l|ccc|c}
\toprule
\textbf{Method}  & \makecell{\textbf{Grasp Single} \\ \textbf{Opened Coke Can}} & 
\makecell{\textbf{Move Near} \\ \textbf{Google Baked Tex}} & 
\makecell{\textbf{Close/Open} \\ \textbf{Drawer Custom}} & \textbf{Avg.}\\
\midrule
 Full-Training &  95.8 & 87.5 & 41.7 & 75.0 \\
Random & 100.0 & 66.7 & 0.0  & 55.6 \\
DemInf & 100.0 & 85.4 & 16.7  & 67.4 \\
SCIZOR & 99.0 & 87.5 & 29.2  & 71.9 \\
\rowcolor{navyblue!9}SIEVE (Ours)  & 100.0 & 91.7 & 37.5  & 76.4 \\
\bottomrule
\end{tabular}
}
\caption{Detailed results on Fractal. This table expands the Fractal results reported in Table~\ref{tab:exp_diff_datasets}, showing the success rate for each evaluation task. Full-Training uses the complete dataset with 100K training steps, while all data selection methods use 50\% of the training data and 50\% of the training steps.}
\label{tab:app_fractal}
\end{table}

\begin{table}[!tb]
\centering
\renewcommand{\arraystretch}{1.2}
\setlength\tabcolsep{4pt} 
\resizebox{\linewidth}{!}
{
\begin{tabular}{l|ccccc|c}
\toprule
\textbf{Method}  & \makecell{\textbf{PnP *} \\ \textbf{To * Close}} & 
\makecell{\textbf{PnP Novel From} \\ \textbf{Cuttingboard To *}} & 
\makecell{\textbf{PnP Novel From} \\ \textbf{Placemat To *}} &  
\makecell{\textbf{PnP Novel From} \\ \textbf{Plate To *}} &
\makecell{\textbf{PnP Novel From} \\ \textbf{Tray To *}} & \textbf{Avg.}\\
\midrule
 Full-Training & 52.8 &  54.8 &  46.5 &  55.5 & 54.0 & 52.7  \\
Random & 50.5 &  56.8 &  49.8 &  60.3 &  50.1 & 53.5 \\
DemInf & 53.3 &  55.1 &  48.6 &  58.1 &  53.9 & 53.8 \\
SCIZOR & 54.0 & 56.1 & 48.1 &  60.5 & 52.3 & 54.2 \\
\rowcolor{navyblue!9}SIEVE (Ours) & 55.3 & 56.7 & 47.9 &  61.3 & 52.8 & 54.8 \\
\bottomrule
\end{tabular}
}
\caption{Detailed results on GR00T-X-Sim. This table expands the GR00T-X-Sim results reported in Table~\ref{tab:exp_diff_datasets}, showing the average success rate for each RoboCasa-GR1 task category. Full-Training uses the complete dataset with 100K training steps, while all data selection methods use 50\% of the training data and 50\% of the training steps.}
\label{tab:app_gr00t}
\end{table}

%% file: aaai2026.bib
@inproceedings{zitkovich2023rt,
  title={Rt-2: Vision-language-action models transfer web knowledge to robotic control},
  author={Zitkovich, Brianna and Yu, Tianhe and Xu, Sichun and Xu, Peng and Xiao, Ted and Xia, Fei and Wu, Jialin and Wohlhart, Paul and Welker, Stefan and Wahid, Ayzaan and others},
  booktitle={Conference on Robot Learning},
  pages={2165--2183},
  year={2023},
  organization={PMLR}
}

@inproceedings{o2024open,
  title={Open x-embodiment: Robotic learning datasets and rt-x models: Open x-embodiment collaboration 0},
  author={O’Neill, Abby and Rehman, Abdul and Maddukuri, Abhiram and Gupta, Abhishek and Padalkar, Abhishek and Lee, Abraham and Pooley, Acorn and Gupta, Agrim and Mandlekar, Ajay and Jain, Ajinkya and others},
  booktitle={2024 IEEE International Conference on Robotics and Automation (ICRA)},
  pages={6892--6903},
  year={2024},
  organization={IEEE}
}

@article{kim2024openvla,
  title={Openvla: An open-source vision-language-action model},
  author={Kim, Moo Jin and Pertsch, Karl and Karamcheti, Siddharth and Xiao, Ted and Balakrishna, Ashwin and Nair, Suraj and Rafailov, Rafael and Foster, Ethan and Lam, Grace and Sanketi, Pannag and others},
  journal={arXiv preprint arXiv:2406.09246},
  year={2024}
}

@article{chi2025diffusion,
  title={Diffusion policy: Visuomotor policy learning via action diffusion},
  author={Chi, Cheng and Xu, Zhenjia and Feng, Siyuan and Cousineau, Eric and Du, Yilun and Burchfiel, Benjamin and Tedrake, Russ and Song, Shuran},
  journal={The International Journal of Robotics Research},
  volume={44},
  number={10-11},
  pages={1684--1704},
  year={2025},
  publisher={Sage Publications Sage UK: London, England}
}

@article{black2024pi_0,
  title={$\pi_0$: A Vision-Language-Action Flow Model for General Robot Control},
  author={Black, Kevin and Brown, Noah and Driess, Danny and Esmail, Adnan and Equi, Michael and Finn, Chelsea and Fusai, Niccolo and Groom, Lachy and Hausman, Karol and Ichter, Brian and others},
  journal={arXiv preprint arXiv:2410.24164},
  year={2024}
}

@article{intelligence2025pi_,
  title={$\pi_{0.5}$: a Vision-Language-Action Model with Open-World Generalization},
  author={Intelligence, Physical and Black, Kevin and Brown, Noah and Darpinian, James and Dhabalia, Karan and Driess, Danny and Esmail, Adnan and Equi, Michael and Finn, Chelsea and Fusai, Niccolo and others},
  journal={arXiv preprint arXiv:2504.16054},
  year={2025}
}

@article{intelligence2026pi,
  title={$\pi_{0.7}$: a Steerable Generalist Robotic Foundation Model with Emergent Capabilities},
  author={Intelligence, Physical and Ai, Bo and Amin, Ali and Aniceto, Raichelle and Balakrishna, Ashwin and Balke, Greg and Black, Kevin and Bokinsky, George and Cao, Shihao and Charbonnier, Thomas and others},
  journal={arXiv preprint arXiv:2604.15483},
  year={2026}
}

@article{kim2025fine,
  title={Fine-tuning vision-language-action models: Optimizing speed and success},
  author={Kim, Moo Jin and Finn, Chelsea and Liang, Percy},
  journal={arXiv preprint arXiv:2502.19645},
  year={2025}
}

@article{bjorck2025gr00t,
  title={Gr00t n1: An open foundation model for generalist humanoid robots},
  author={Bjorck, Johan and Casta{\~n}eda, Fernando and Cherniadev, Nikita and Da, Xingye and Ding, Runyu and Fan, Linxi and Fang, Yu and Fox, Dieter and Hu, Fengyuan and Huang, Spencer and others},
  journal={arXiv preprint arXiv:2503.14734},
  year={2025}
}

@article{xing2025shortcut,
  title={Shortcut learning in generalist robot policies: The role of dataset diversity and fragmentation},
  author={Xing, Youguang and Luo, Xu and Xie, Junlin and Gao, Lianli and Shen, Hengtao and Song, Jingkuan},
  journal={arXiv preprint arXiv:2508.06426},
  year={2025}
}

@article{lian2026intentvla,
  title={IntentVLA: Short-Horizon Intent Modeling for Aliased Robot Manipulation},
  author={Lian, Shijie and Yu, Bin and Lin, Xiaopeng and Shen, Zhaolong and Yang, Laurence Tianruo and Jin, Yurun and Liu, Haishan and Wu, Changti and Yuan, Hang and Huang, Cong and others},
  journal={arXiv preprint arXiv:2605.14712},
  year={2026}
}

@article{hussein2017imitation,
  title={Imitation learning: A survey of learning methods},
  author={Hussein, Ahmed and Gaber, Mohamed Medhat and Elyan, Eyad and Jayne, Chrisina},
  journal={ACM Computing Surveys (CSUR)},
  volume={50},
  number={2},
  pages={1--35},
  year={2017},
  publisher={ACM New York, NY, USA}
}

@inproceedings{ross2010efficient,
  title={Efficient reductions for imitation learning},
  author={Ross, St{\'e}phane and Bagnell, Drew},
  booktitle={Proceedings of the thirteenth international conference on artificial intelligence and statistics},
  pages={661--668},
  year={2010},
  organization={JMLR Workshop and Conference Proceedings}
}

@article{sathyanarayan2025quality,
  title={Quality Over Quantity: Curating Contact-Based Robot Datasets Improves Learning},
  author={Sathyanarayan, Hrishikesh and Vantilborgh, Victor and Abraham, Ian},
  journal={arXiv preprint arXiv:2510.18137},
  year={2025}
}

@article{belkhale2023data,
  title={Data quality in imitation learning},
  author={Belkhale, Suneel and Cui, Yuchen and Sadigh, Dorsa},
  journal={Advances in neural information processing systems},
  volume={36},
  pages={80375--80395},
  year={2023}
}

@inproceedings{lin2025data,
  title={Data scaling laws in imitation learning for robotic manipulation},
  author={Lin, Fanqi and Hu, Yingdong and Sheng, Pingyue and Wen, Chuan and You, Jiacheng and Gao, Yang},
  booktitle={International Conference on Learning Representations},
  volume={2025},
  pages={54877--54910},
  year={2025}
}

@article{xiao2025data,
  title={Data Assessment for Embodied Intelligence},
  author={Xiao, Jiahao and Yan, Bowen and Zhang, Jianbo and Wang, Jia and Li, Chunyi and Cheng, Zhengxue and Zhai, Guangtao},
  journal={arXiv preprint arXiv:2511.09119},
  year={2025}
}

@article{hejna2024re,
  title={Re-mix: Optimizing data mixtures for large scale imitation learning},
  author={Hejna, Joey and Bhateja, Chethan and Jiang, Yichen and Pertsch, Karl and Sadigh, Dorsa},
  journal={arXiv preprint arXiv:2408.14037},
  year={2024}
}

@article{zhang2025scizor,
  title={Scizor: A self-supervised approach to data curation for large-scale imitation learning},
  author={Zhang, Yu and Xie, Yuqi and Liu, Huihan and Shah, Rutav and Wan, Michael and Fan, Linxi and Zhu, Yuke},
  journal={arXiv preprint arXiv:2505.22626},
  year={2025}
}

@article{hejna2025robot,
  title={Robot data curation with mutual information estimators},
  author={Hejna, Joey and Mirchandani, Suvir and Balakrishna, Ashwin and Xie, Annie and Wahid, Ayzaan and Tompson, Jonathan and Sanketi, Pannag and Shah, Dhruv and Devin, Coline and Sadigh, Dorsa},
  journal={arXiv preprint arXiv:2502.08623},
  year={2025}
}

@article{dass2025datamil,
  title={Datamil: Selecting data for robot imitation learning with datamodels},
  author={Dass, Shivin and Khaddaj, Alaa and Engstrom, Logan and Madry, Aleksander and Ilyas, Andrew and Mart{\'\i}n-Mart{\'\i}n, Roberto},
  journal={arXiv preprint arXiv:2505.09603},
  year={2025}
}

@article{chen2025curating,
  title={Curating demonstrations using online experience},
  author={Chen, Annie S and Lessing, Alec M and Liu, Yuejiang and Finn, Chelsea},
  journal={arXiv preprint arXiv:2503.03707},
  year={2025}
}

@article{yu2026frameskip,
  title={FrameSkip: Learning from Fewer but More Informative Frames in VLA Training},
  author={Yu, Bin and Lian, Shijie and Lin, Xiaopeng and Shen, Zhaolong and Wei, Yuliang and Wu, Changti and Yuan, Hang and Liu, Haishan and Wang, Bailing and Huang, Cong and others},
  journal={arXiv preprint arXiv:2605.13757},
  year={2026}
}

@article{xu2026athena,
  title={ATHENA: Accelerated Multi-Task Heterogeneous Influence Functions for Robot Data Curation},
  author={Xu, Tao and Wang, Jiaxin and Zhang, Runhao and Guan, Jiayi and Zeng, Xianchao and Song, Weixi and Zhou, Xinyu and Chen, Zhetao and Chen, Guang and Li, Yong-Lu},
  journal={arXiv preprint arXiv:2606.16208},
  year={2026}
}

@article{wu2026scalselect,
  title={ScalSelect: Scalable Training-Free Multimodal Data Selection for Efficient Visual Instruction Tuning},
  author={Wu, Changti and Mao, Jiahuai and Miao, Yuzhuo and Lian, Shijie and Yu, Bin and Lin, Xiaopeng and Huang, Cong and Zhang, Lei and Chen, Kai},
  journal={arXiv preprint arXiv:2602.11636},
  year={2026}
}

@article{zhou2026synthetic,
  title={Synthetic Data for Multimodal Large Language Models: A Lifecycle-Oriented Survey},
  author={Zhou, Yue and Chang, Yi and Wu, Yuan},
  year={2026},
  publisher={Preprints}
}

@article{rissanen2004minimum,
  title={Minimum description length principle},
  author={Rissanen, Jorma},
  journal={Encyclopedia of statistical sciences},
  volume={7},
  year={2004},
  publisher={Wiley Online Library}
}

@book{grunwald2007minimum,
  title={The minimum description length principle},
  author={Gr{\"u}nwald, Peter D},
  year={2007},
  publisher={MIT press}
}

@article{barron1998minimum,
  title={The minimum description length principle in coding and modeling},
  author={Barron, Andrew and Rissanen, Jorma and Yu, Bin},
  journal={IEEE transactions on information theory},
  volume={44},
  number={6},
  pages={2743--2760},
  year={1998},
  publisher={IEEE}
}

@article{finzi2026entropy,
  title={From Entropy to Epiplexity: Rethinking Information for Computationally Bounded Intelligence},
  author={Finzi, Marc and Qiu, Shikai and Jiang, Yiding and Izmailov, Pavel and Kolter, J Zico and Wilson, Andrew Gordon},
  journal={arXiv preprint arXiv:2601.03220},
  year={2026}
}

@article{assran2025v,
  title={V-jepa2: Self-supervised video models enable understanding, prediction and planning},
  author={Assran, Mido and Bardes, Adrien and Fan, David and Garrido, Quentin and Howes, Russell and Muckley, Matthew and Rizvi, Ammar and Roberts, Claire and Sinha, Koustuv and Zholus, Artem and others},
  journal={arXiv preprint arXiv:2506.09985},
  year={2025}
}

@inproceedings{sculley2010web,
  title={Web-scale k-means clustering},
  author={Sculley, David},
  booktitle={Proceedings of the 19th international conference on World wide web},
  pages={1177--1178},
  year={2010}
}

@inproceedings{walke2023bridgedata,
  title={Bridgedata v2: A dataset for robot learning at scale},
  author={Walke, Homer Rich and Black, Kevin and Zhao, Tony Z and Vuong, Quan and Zheng, Chongyi and Hansen-Estruch, Philippe and He, Andre Wang and Myers, Vivek and Kim, Moo Jin and Du, Max and others},
  booktitle={Conference on Robot Learning},
  pages={1723--1736},
  year={2023},
  organization={PMLR}
}

@article{li2024evaluating,
  title={Evaluating real-world robot manipulation policies in simulation},
  author={Li, Xuanlin and Hsu, Kyle and Gu, Jiayuan and Pertsch, Karl and Mees, Oier and Walke, Homer Rich and Fu, Chuyuan and Lunawat, Ishikaa and Sieh, Isabel and Kirmani, Sean and others},
  journal={arXiv preprint arXiv:2405.05941},
  year={2024}
}

@article{nasiriany2024robocasa,
  title={Robocasa: Large-scale simulation of everyday tasks for generalist robots},
  author={Nasiriany, Soroush and Maddukuri, Abhiram and Zhang, Lance and Parikh, Adeet and Lo, Aaron and Joshi, Abhishek and Mandlekar, Ajay and Zhu, Yuke},
  journal={arXiv preprint arXiv:2406.02523},
  year={2024}
}

@article{bai2025qwen3,
  title={Qwen3-vl technical report},
  author={Bai, Shuai and Cai, Yuxuan and Chen, Ruizhe and Chen, Keqin and Chen, Xionghui and Cheng, Zesen and Deng, Lianghao and Ding, Wei and Gao, Chang and Ge, Chunjiang and others},
  journal={arXiv preprint arXiv:2511.21631},
  year={2025}
}

@article{lin2025physbrain,
  title={Physbrain: Human egocentric data as a bridge from vision language models to physical intelligence},
  author={Lin, Xiaopeng and Lian, Shijie and Yu, Bin and Yang, Ruoqi and Shen, Zhaolong and Wu, Changti and Miao, Yuzhuo and Jin, Yurun and Shi, Yukun and He, Jiyan and others},
  journal={arXiv preprint arXiv:2512.16793},
  year={2025}
}

@article{shao2025large,
  title={Large vlm-based vision-language-action models for robotic manipulation: A survey},
  author={Shao, Rui and Li, Wei and Zhang, Lingsen and Zhang, Renshan and Liu, Zhiyang and Chen, Ran and Nie, Liqiang},
  journal={arXiv preprint arXiv:2508.13073},
  year={2025}
}
